\UseRawInputEncoding 
\documentclass{egpubl}

\JournalSubmission    
\usepackage[T1]{fontenc}
\usepackage{dfadobe}  

\usepackage{cite}  
\BibtexOrBiblatex
\electronicVersion
\PrintedOrElectronic
\ifpdf \usepackage[pdftex]{graphicx} \pdfcompresslevel=9
\else \usepackage[dvips]{graphicx} \fi

\usepackage{egweblnk} 
\usepackage{arydshln}

\usepackage{amssymb}
\usepackage{amsmath}
\usepackage{relsize}

\title[SketchZooms]%
{SketchZooms: Deep Multi-view Descriptors for Matching Line Drawings}
\author[P.~Navarro, J.~I.~Orlando, C.~Delrieux \&  E.~Iarussi]
{P.~Navarro$^{1,2,3}$\orcid{0000-0003-2180-449X}, \ J.~I.~Orlando$^{3,4}$\orcid{0000-0001-9734-5571}, \ 
C.~Delrieux$^{3,5}$\orcid{0000-0002-2727-8374}, and \ 
E.~Iarussi$^{3,6}$\orcid{0000-0001-7438-9299} \\
	        $^1$ Instituto Patagónico de Ciencias Sociales y Humanas, CENPAT, Puerto Madryn, Argentina\\
	        $^2$ Departamento de Informática, Universidad Nacional de la Patagonia San Juan Bosco, Trelew, Argentina\\
	        $^3$Consejo Nacional de Investigaciones Científicas y Técnicas (CONICET), Argentina\\
			$^4$Instituto Pladema, UNICEN, Tandil, Argentina\\
			$^5$Departamento de Ing. Eléctrica y Computadoras, Universidad Nacional del Sur (UNS), Bahia Blanca, Argentina\\
			$^6$Universidad Tecnológica Nacional (UTN FRBA), Buenos Aires, Argentina
	}

\begin{document}


\maketitle
\begin{abstract}
Finding point-wise correspondences between images is a long-standing problem in image analysis.
This becomes particularly challenging for sketch images, due to the varying nature of human drawing style, projection distortions and viewport changes. 
In this paper we present the first attempt to obtain a learned descriptor for dense registration in line drawings.
Based on recent deep learning techniques for corresponding photographs, we designed descriptors to locally match image pairs where the object of interest belongs to the same semantic category, yet still differ drastically in shape, form, and projection angle. 
To this end, we have specifically crafted a data set of synthetic sketches using non-photorealistic rendering over a large collection of part-based registered 3D models. 
After training, a neural network generates descriptors for every pixel in an input image, which are shown to generalize correctly in unseen sketches hand-drawn by humans.  
We evaluate our method against a baseline of correspondences data collected from expert designers, in addition to comparisons with other descriptors that have been proven effective in sketches. 
Code, data and further resources will be publicly released by the time of publication.



\end{abstract}

\section{Introduction}

\begin{figure}[!b]
    \centering
    \includegraphics[width=1.0\linewidth]{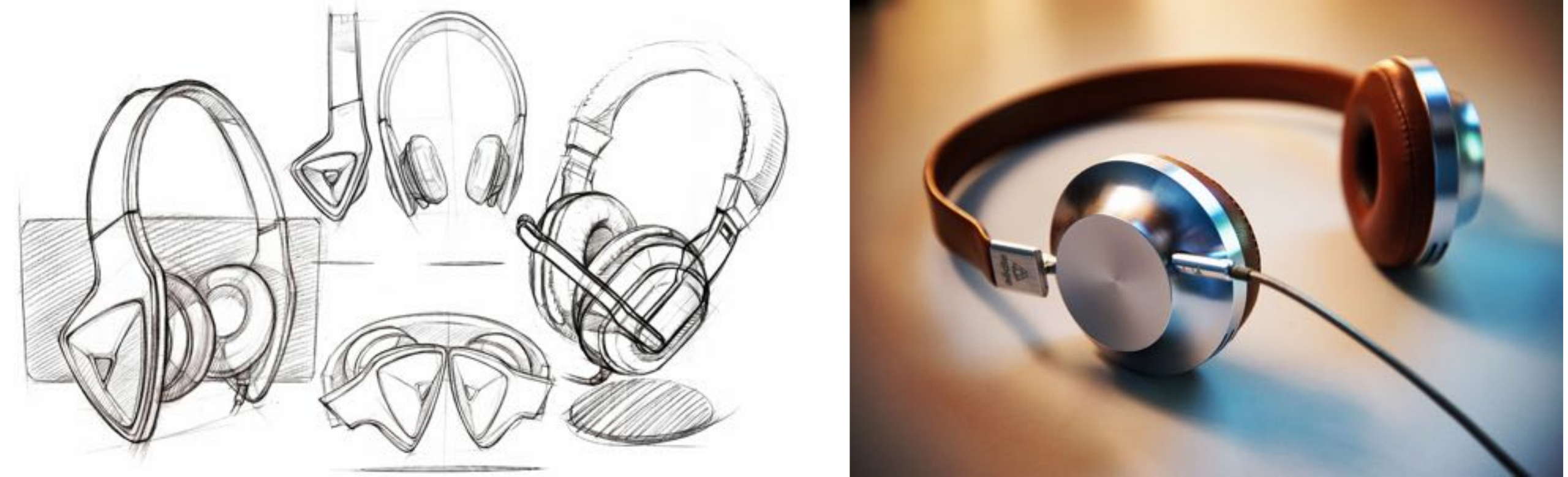}
    \parbox[t]{\columnwidth}{\relax }
    \caption{Unlike photographs, typical design sketches lack shading, texture, and lines are often rough and incomplete.
    \label{fig:fig_1_concept}}
\end{figure}

Humans excel at perceiving 3D objects from line drawings \cite{hertzmann2020line}. 
Therefore, freehand sketches are still the preferred way for artists and designers to express and communicate shape without requiring to construct the intended object. 
Unlike humans, computers struggle to interpret a 2D sketch as a highly condensed abstraction of our 3D world. 
For instance, the straightforward task of finding correspondences between a pair of images or an image and a 3D model has been an important problem in Computer Graphics and Vision for decades.
In comparison with photographs, dealing with sketches is even more challenging \cite{arora2017sketchsoup}, since line drawings lack key shape cues like shading and texture, projections are imprecise, and shapes are often composed by several sketchy lines (Figure~\ref{fig:fig_1_concept}). 
Consequently, when a target object is viewed from different angles, traditional image descriptors fail to map similar points close together in the descriptor space.
Furthermore, recent studies show that even advanced deep networks lack the ability to generalise to sketches when originally trained to perform perceptual tasks over photo collections \cite{lamb2019sketchtransfer}. 

\begin{figure}[!t]
  \includegraphics[width=1\linewidth]{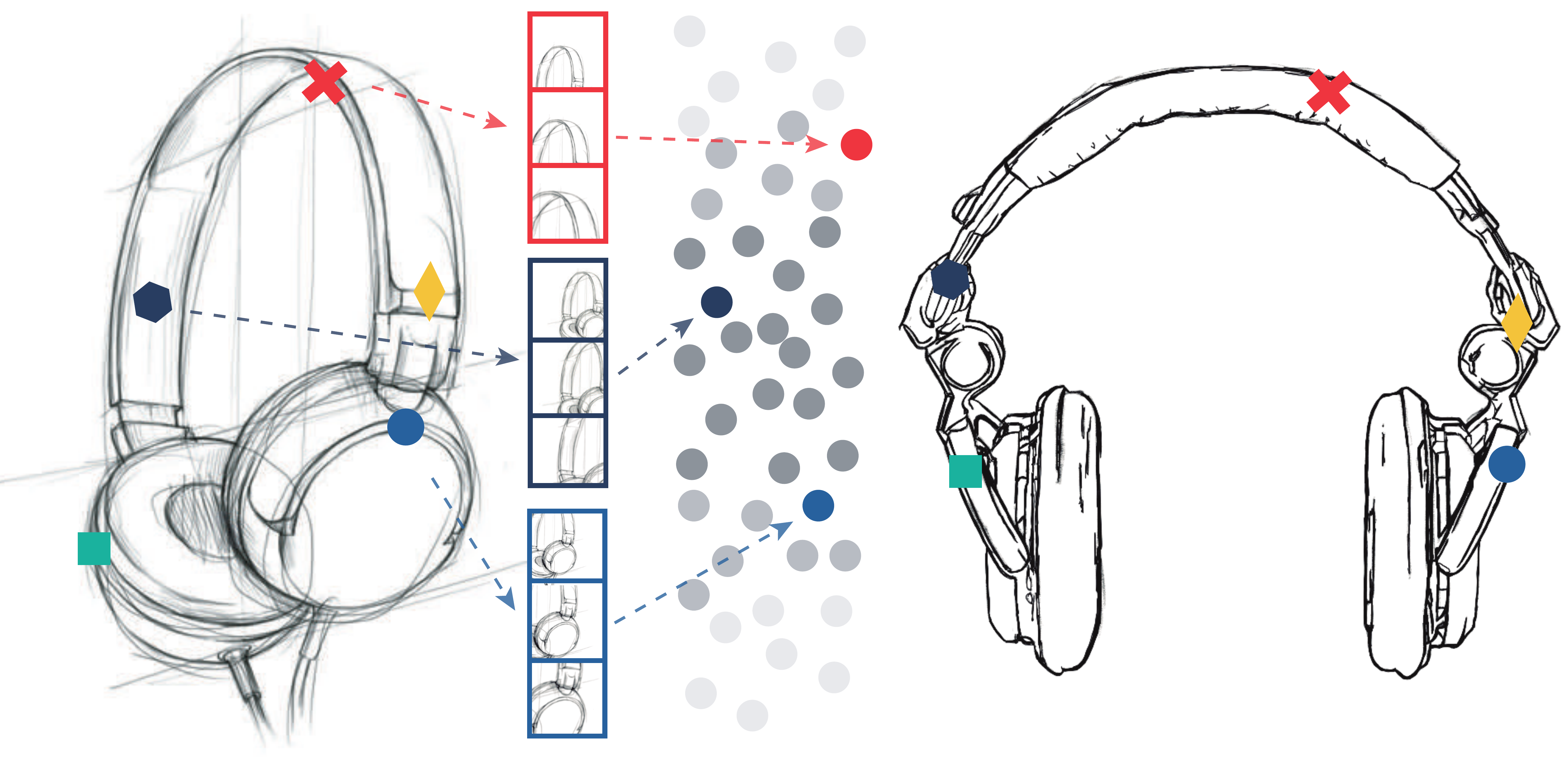}
   \caption{The multi-view neural network embeds similar points on sketches close to one another in descriptor space despite the significant changes in viewport and shape. By training on a dataset of part-based registered 3D models rendered as sketches, SketchZooms is able to generalise to different rendering styles, incorporating the semantics from an object category.}
   \label{fig:fig_2_teaser}
\end{figure}

To date, finding local sketch correspondences with deep learning techniques is an unexplored research topic.
This is likely because learning meaningful and consistent features using such high capacity models requires a large dataset of complex line drawings, paired semantically at a dense, pixel-wise level.
To overcome this difficulty, our key contribution is a vast collection of synthetic sketches, distributed in several semantic categories.
This massive dataset serves to compute local sketch descriptors that can deal with significant image changes.
In our setup, a query point is represented by a set of 2D zooms captured from the point's immediate neighbourhood, resulting in multiple zoomed versions of the corresponding point. 
The main goal is to capture the domain semantics and object part characteristics despite the heterogeneous nature of hand-drawn images (see Figure~\ref{fig:fig_2_teaser}).
Our hypothesis is that learning from such a large database may result in a general model that overcomes the covariate shift between artificial and real sketches. 

Although the available literature provides descriptors which are robust to shape variation and affine distortions, to the best of our knowledge this is the first attempt to cope with part semantics and 3D viewport changes in sketches. 
To this avail, we evaluate and compare SketchZooms against well-known state-of-the-art techniques extensively used in line drawings' applications.
Furthermore, the generalisation ability of our framework is assessed by evaluating the proposed approach using the OpenSketch image collection, \cite{gryaditskaya2019opensketch} rendered by designers in different styles and viewports.
Taking advantage of co-registered images in OpenSketch, without any prior fine-tuning, we compute quantitative metrics along with the qualitative results (Section~\ref{sec:results}).
Experiments show that this approach is able to deal with significant changes in style, shape, and viewport, generalising well to non-synthetic inputs.
Finally, we demonstrate the usefulness of our descriptors for graphics applications such as sketch-based shape retrieval, and image morphing.

In summary, our contributions are:
\begin{itemize}
    \item The first approach applying deep learning to the problem of finding local image correspondences between design sketches. 
    \item A massive co-registered synthetic line drawing dataset rendered from 3D shapes, which allows our trained models to generalise to designer sketches, even from unseen object categories.
    \item A comprehensive evaluation and comparison with respect to other methods commonly applied to find correspondences in line drawings, including a perceptual study against human-established matchings.
\end{itemize}

\section{Related Work}\label{sec:related}

Finding image descriptors that effectively represent image data is a classic problem in Computer Graphics and Vision. 
A comprehensive summary of the relevant literature is out of the scope of this paper. 
Instead, we focus on descriptors involving line drawings, either for registration or retrieval tasks on images and 3D models. 
We avoid general natural image descriptors like SIFT~\cite{lowe1999object} which has been shown ineffective to cope with sparse stroke orientations in sketches~\cite{eitz2012sketch}.
We briefly classify them into two main groups: \emph{hand-crafted} and \emph{learned descriptors}.

\emph{Hand-crafted descriptors} consist on applying custom transformations over some input data in order to obtain a suitable global or local representation.
Many applications working with raster input employ pixel-based descriptors.
For instance, ShapeContext~\cite{belongie2002shape} is a well known descriptor that captures the point distribution on a given neighbourhood, which was proven effective for corresponding feature points in sketches~\cite{chen2009sketch2photo,iarussi2013drawing}. 
Combined with cycle consistency methods like FlowWeb~\cite{zhou2015flowweb}, some authors boosted ShapeContext performance and benefited from the availability of multiple similar sketches \cite{arora2017sketchsoup}.
In the context of vector graphics, several authors proposed to quantify stroke similarity in order to generate in-between frames for character animation~\cite{whited2010betweenit, xing2015autocomplete}, auto-complete line drawings repetitions~\cite{xing2014autocomplete}, selection and grouping~\cite{xu2012lazy,noris2012smart} and sketch beautification~\cite{liu2015closure,liu2018strokeaggregator}.
As the number of available 3D models and images steadily increases, effective methods for searching on databases have emerged. 
Using non-photorealistic rendering methods, meshes are transformed into sketches and search engines compute image descriptors that summarise global properties, such as contour histograms \cite{pu20052d}, stroke similarity distance~\cite{shao2011discriminative}, Fourier transform~\cite{shin2007magic}, diffusion tensor fields~\cite{yoon2010sketch} and bag-of-features models \cite{eitz2012sketch}. 

\emph{Learned descriptors} gained popularity with the recent success of deep neural networks~\cite{lecun2015deep}. 
Most applications involving line drawings, like Sketch Me That Shoe~\cite{yu2016sketch,song2017deep}, target the problem of computing global descriptors for sketch-based image retrieval.
Similarly, Qi~et~al.~\cite{qi2016sketch} and Bui~et~al. \cite{bui2017compact} proposed to train siamese networks that pulls feature vectors closer for sketch-image input pairs labeled as similar, and push them away if irrelevant. 
Zhu~et~al.~\cite{zhu2016learning} constructed pyramid cross-domain neural networks to map sketch and 3D shape low-level representations onto an unified feature space.
Other authors investigated how to learn cross-modal representations that surpass sketch images and 3D shapes, incorporating text labels, descriptions, and even depth maps~\cite{tasse2016shape2vec,castrejon2016learning,zhu2017training}. 
Other learned descriptors applications include sketch classification and recognition~\cite{yu2015sketch,zhang2016sketchnet}.
Like Yu et al. \cite{yu2015sketch}, all these methods target global features that can discriminate high level characteristics in sketches a single representation for an entire shape), our goal is to compute accurate pixel-wise descriptors that capture part semantics along with local and global contexts to perform local matching.

In the context of learning local semantic descriptors for photographs, a common strategy consists on training siamese architectures with pairs or triplets of corresponding and non-corresponding patches. 
Most of these methods~\cite{han2015matchnet,zagoruyko2015learning,simo2015discriminative,choy2016universal,kumar2016learning,tian2017l2} learn representations for natural image patches such that patches depicting the same underlying surface pattern tend to have similar representations. 
In contrast, we aim to learn a deep learning model able to assign similar descriptors to geometrically but also semantically similar points across different objects.
Moreover, instead of a descriptor for a single image patch, our method learns a complex representation for a 3D surface point (depicted as a sketch) by exploiting the information from different views and multiple scales.
Other proposals such as \cite{kim2017fcss} learn a convolutional descriptor using self-similarity, called fully convolutional self-similarity (FCSS), and combine the learned descriptors with the proposal flow framework \cite{ham2016proposal}. 
These approaches to learning semantic correspondences \cite{zhou2016learning} or semantic descriptors \cite{han2017scnet} generally perform better than traditional hand-crafted ones.
However, since limited training data is available for semantic correspondence in photographs, these region-based methods rely on weakly-supervised feature learning schemes, leveraging correspondence consistency between object locations provided in existing image data sets. 
This makes them vulnerable to changes in orientation and projection distortion, and also to shape variation as commonly seen in line drawings, where the number and style of strokes may change significantly while the semantics of the parts are preserved.

Learned descriptors require adequate training datasets. 
The high diversity in style and the difficulty to automate sketch annotation makes it hard to compile massive line drawing datasets.
Eitz~et~al.~\cite{eitz2012humans} introduced a dataset of 20,000 sketches spanning 250 categories. 
Similarly, The Sketchy Database~\cite{sangkloy2016sketchy} ask crowd workers to sketch photographic objects sampled from 125 categories and acquired 75,471 sketches, compiling the first large-scale collection of sketch-photo pairs. 
Recently, Quick, Draw!~\cite{ha2017neural} released an open source collection composed by 50 million doodles across 345 categories drawn by players of an online game. 
Nevertheless, the skills and style disparities of contributors to these datasets makes them unsuitable for our goal. 
Instead, we target design sketches that are drawn following approximately a particular set of rules~\cite{eissen2011sketching}. 
Similar to Wang~et~al.~\cite{wang2015sketch}, we exploit shape collections augmented with semantic part-based correspondences data to synthesise sketches with NPR techniques. 
The registered 3D models naturally provide us with 2D/3D alignment, a crucial ingredient to learn our multidimensional features.

\section{Multi-view Sketch Data}\label{sec:dataset}

\emph{Shape collection.} 
As with recent work targeting sketches and machine learning~\cite{delanoy20173d,huang2016shape,su2018interactive}, we generated synthetic line drawings based on semantically corresponded 3D shapes.  
From the ShapeNetCore dataset \cite{yi2016scalable} we selected models in 16 categories: airplane (1,000), bag (152), cap (110), car (1,000), chair (1,000), earphone (138), guitar (1,000), knife (784), lamp (1,000), laptop (890), motorbike (404), mug (368), pistol (550), rocket (132), skateboard (304), and table (1,000).
The 3D models were augmented with correspondences files that provide a list of 10,000 randomly sampled surface matching points for every possible pair of shapes within each category. 
Correspondences are computed with a part-based registration algorithm that performs a non-rigid alignment of all pairs of segments with the same label over two target shapes as proposed in \cite{huang2018learning}. 

\begin{figure}[!t] 
    \centering
    \includegraphics[width=1.0\linewidth]{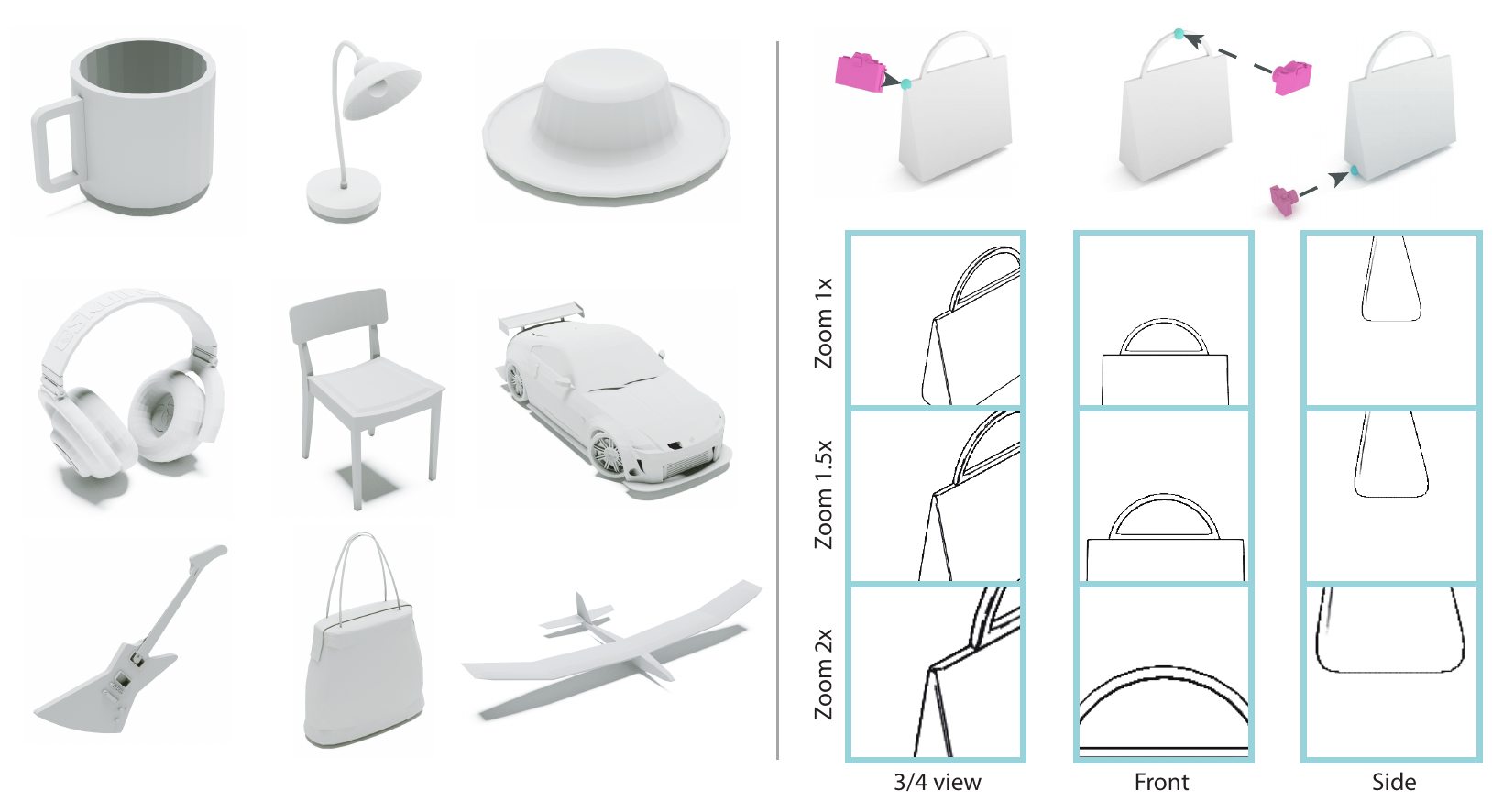}
    \parbox[t]{\columnwidth}{ }
    \caption{Visualisation of images in our dataset. Given a collection of almost 10,000 3D models distributed in 16 object categories, we rendered line drawings from three predefined angles and three different distances to the target surface point.}
    \label{fig:fig_3_dataset}
\end{figure}

\emph{Synthetic line drawings}. 
While previous data-driven methods in sketching employ simple models such as Canny edges [1987] or image-space contours from~\cite{saito1990comprehensible}, we adopted Apparent Ridges from~\cite{judd2007apparent}, a good approximation to artists lines as shown in~{\em Where do people draw lines?}~\cite{cole2012people}.
Apparent Ridges' lines approximate meaningful shape cues commonly drawn by humans to convey 3D objects.
Apart from rendering style, viewport selection is crucial to convey shape in sketches.  
Literature in design recommends to adopt specific viewports in order to reduce the sketch ambiguity and simultaneously show most of the target shape~\cite{eissen2011sketching}. 
Most 3D reconstruction algorithms from sketch images rely on assumptions like parallelism, orthogonality, and symmetry~\cite{cordier2016sketch}.
Following these guidelines, we selected a set of orthographic views to  render from 3D models. 
We used a total of two accidental (object-aligned) views: front, right side, and a single isometric angle, also called informative view: front-right (see Figure~\ref{fig:fig_3_dataset}). 
Left sides were omitted since we assume that the objects are symmetric with respect to the front view (see Section~\ref{sec:limit}). 
To capture images, we centred an orthographic camera on each sample point, and shoot it from three different constant distances and three distinct viewports (successive zooms at 1.0x, 1.5x, and 2x).
Occluded points were discarded by comparing z-buffer data with camera-to-target point distance. 
Rendering all the sampled surface points for each model would have been very computationally expensive. 
On the other hand, discarding some models from each category would not have been a good option, since the diversity of shapes favours generalisation. 
For these reasons, we decided to randomly sub-sample surface points on each model, so that we only use a fraction of them.
For every model in our dataset, we randomly choose and render 70 corresponding points to other models within its category.
To increase diversity, we randomly select the sub-sampled points differently for each model.
In total, our dataset consists of 538,000 images in a resolution of 512$\times$512 pixels.
Each image has been augmented with the information needed to retrieve all other corresponding points in the dataset.
It took approximately 15 days to complete the rendering stage on a PC equipped with an NVIDIA Titan Xp GPU and an Intel i7 processor. 
For the sake of reproducibility and to encourage further research, we aim to publicly release it by the time of publication.

\section{Learning Multi-view Descriptors for Line Drawings}\label{sec:approach}

\subsection{Proposed Approach}

\emph{Backbone architecture.} Our proposed approach relies on a backbone convolutional neural network (CNN) that is responsible for learning and computing descriptors for each given input. We choose a CNN inspired by the standard AlexNet~\cite{krizhevsky2012imagenet}, which comprises five convolutional layers, followed by ReLU non-linearities and max pooling (see supplemental material for details).
Nevertheless, our method is sufficiently general to incorporate any other backbone architecture. 
In Section \ref{sec:abblation} we show SketchZooms performance when using other state of the art networks like VGG19 \cite{simonyan2014very} and ResNet-18 \cite{he2016deep}.
A key insight in our approach consists of aggregating local surface information across multiple zooms.
Therefore, we modified all the aforementioned architectures to incorporate a pooling layer that aggregates the descriptors $Y_{z,p}, z \in Z$ generated for each of the three input zooms $X_{z,p}$ into a single one $Y_{p}=\max\limits_{z} (Y_{z,p})$. 
The aggregation is performed on an element-wise maximum operation across the input zooms. 

\emph{Loss function.} A key component in our approach is the learning mechanism for tuning the network parameters.
We adopted a triplet loss~\cite{schroff2015facenet} motivated by the fact that distances gain richer semantics when put into context, and the anchor point added by the triplet loss better shapes the embedding by exploiting this relativistic approach \cite{wu2017sampling}.
We strive for an embedding from a set of sketch image zooms $X_{z,p}$ centred on a point $p$, into a descriptor $Y_{p} \in \mathbb{R}^{d}$ ($d=128$ in our setup).
Triplet loss minimises the distance between an anchor $Y^{a}$ and a corresponding (also called positive) point descriptor $Y^{c}$.
Simultaneously, it maximises the distance between the anchor and a non-corresponding (negative) point descriptor $Y^{n}$.
Formally, we want: 
\begin{equation} \label{eq:ineq}
     D^{2}\left(Y^{a},Y^{c}\right) + \alpha < D^{2}\left( Y^{a},Y^{n}\right),
\end{equation}
where $D$ stands for the Euclidean distance between descriptors, and 
$\alpha$ is a margin enforced between positive and negative pairs ($\alpha=1$ in our implementation).
Formulating Equation~\ref{eq:ineq} as an optimisation problem over the network parameters $\boldsymbol{w}$, we have: 
\begin{equation} \label{eq:loss}
     \mathcal{L}(\boldsymbol{w}) = \sum_{i}^{N} \max\left(D^{2}\left(Y^{a}_{i},Y^{c}_{i}\right) - D^{2}\left( Y^{a}_{i},Y^{n}_{i}\right) + \alpha,0\right),
\end{equation}
where $N$ is the cardinality of the triplets training set. 

Naively using all triplets is highly inefficient since the more the training progresses, the more triplets are going to satisfy Equation~\ref{eq:ineq}, making training slower over time.
Therefore, we adopted an alternative approach in which we adaptively select semi-hard triplets on each training step satisfying:
\begin{equation} \label{eq:semihard}
   \begin{cases}
        D^{2}\left(Y^{a},Y^{c}\right) < D^{2}\left(Y^{a},Y^{n}\right), \\
        D^{2}\left(Y^{a},Y^{n}\right) < D^{2}\left(Y^{a},Y^{c}\right) + \alpha,
    \end{cases}
\end{equation}
meaning that we look for training samples $\{Y^{a},Y^{c},Y^{n}\}$ lying inside the semi-hard margin area delimited by $\alpha$.
For the sake of notation, we refer to triplets using descriptor notation symbol $Y$.  
In practice, we compute $\{Y^{a},Y^{c},Y^{n}\}$ from input images using the last network training state.
We build useful triplets on the fly for each training minibatch by testing whether their descriptors infringe Equation~\ref{eq:semihard}. 
In our setup, we cluster individual samples in groups $G$ to be sequentially used during each training epoch.  
To build a minibatch, we randomly sample positive pairs from $G$ of the form $[Y^{a}_{i},Y^{c}_{j}],\: i,j \in G $. 
From construction, our data set allows to easily obtain these corresponding pairs since they are exhaustively listed in custom files.
We then test the semi-hard conditions over a random number $s$ of negative samples $[Y^{a}_{i},Y^{n}_{k}],\: i,k, \in G$.
We experimented with several values for $s$ and found $s=5$ to minimize the time spent in random search while still providing good triplets for training.  

\subsection{Experimental Setup}

We experimentally evaluated multiple aspects of our approach: (i) we tested SketchZooms on a number of hand-drawn images from the OpenSketch dataset \cite{gryaditskaya2019opensketch} to assess the generalisation power of the network to unseen shape categories and styles, (ii) we examined the ability of our learned embeddings to properly distribute descriptors in the feature space, (iii) we computed correspondence accuracy metrics to evaluate matching performance in the image space, (iv) we performed a perceptual study to assess the semantic aspects of our features, and (v) we compare the performance on the aforementioned metrics against other correspondences methods, with emphasis on those commonly applied to line drawings. 

\emph{Metrics.} We report quantitative results using two standard metrics.
First, we tested our embedding space using cumulative match characteristic (CMC), a standard quality measure for image correspondences~\cite{karanam2015person, wagg2004automated}. 
This metric captures the proximity between points inside the embedding space by computing distances over descriptor pairs on two target sketches: given a point on one of the input images, a list of corresponding candidate matchings on the other image is retrieved; then, candidates are ranked using a proximity measure, e.g. the Euclidean distance in descriptor space.
We also evaluated the accuracy of our descriptors on the image space using the correspondence accuracy (CAcc) from~\cite{kim2013learning}, over our set of test samples. 
This metric evaluates the accuracy of predicted correspondences with respect to the ground truth by registering all L2 distances between retrieved matching points and ground truths. 
We report the percentage of matchings below normalised euclidean distance (5\% of image side (512 pixels)). 

\begin{figure*}[!t]
    \centering
    \includegraphics[width=1.0\linewidth]{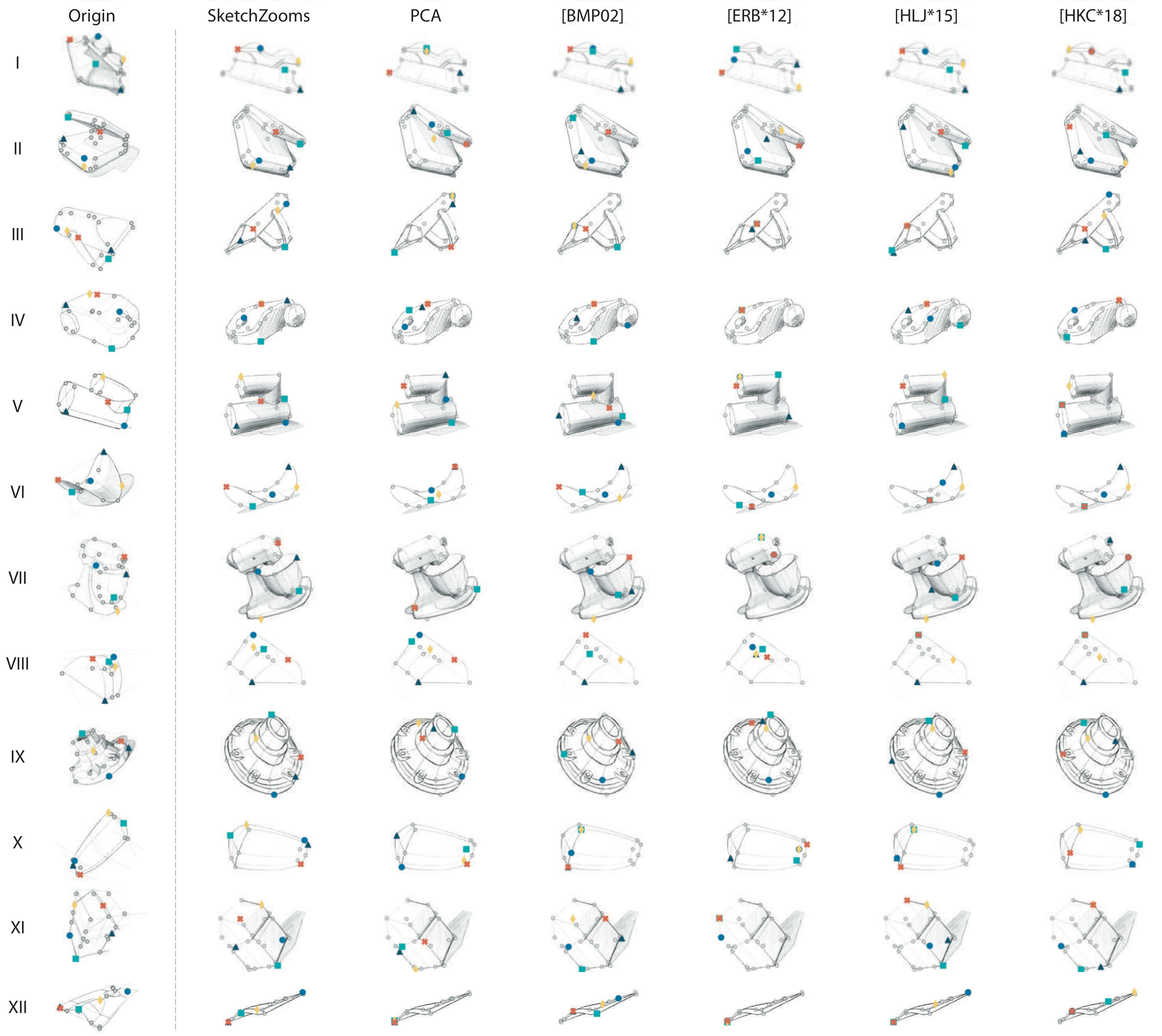}
    \parbox[t]{\columnwidth}{\relax }
    \caption{Pair-wise sparse correspondences on OpenSketch data. Images have been randomly rotated between $\pm$90$^{\circ}$ before computing the descriptors. Columns show origin and target images, corresponded using different local descriptors. For each image pair, we highlighted five points distributed in different areas of the image. The grey dots show the remaining sampled points for which matches were obtained in order to compute metrics in Table \ref{tab:cmc-score}. Only Euclidean distance in feature space has been considered to determine these correspondences. Overall, our learned descriptors manage to identify similar underlying local shapes, despite the extreme differences in style, hatching, shadows, construction lines, and camera positions.}
    \label{fig:fig_4_sparse}
\end{figure*}

\begin{table*}[!t]
\centering
\caption{Cumulative match characteristic (percentage of correct matches obtained in the top 5 rank), and correspondence accuracy (percentage of matchings below 5\%  of the image width in Euclidean space w.r.t. ground truth) on all evaluated methods over OpenSketch dataset. In total, our test samples consist of 66,320 corresponding points.}
\label{tab:cmc-score}
\resizebox{\textwidth}{!}{%
\begin{tabular}{l l l l l l l l l l l l l }
\hline
                                   & \multicolumn{2}{c}{mixer}                           & \multicolumn{2}{c}{tubes}                           & \multicolumn{2}{c}{wobble surface}                  & \multicolumn{2}{c}{hairdryer}                       & \multicolumn{2}{c}{vacuum cleaner}                  & \multicolumn{2}{c}{mouse}                           \\ \hline
                                   & \multicolumn{1}{c}{CMC} & \multicolumn{1}{c}{CAcc} & \multicolumn{1}{c}{CMC} & \multicolumn{1}{c}{CAcc} & \multicolumn{1}{c}{CMC} & \multicolumn{1}{c}{CAcc} & \multicolumn{1}{c}{CMC} & \multicolumn{1}{c}{CAcc} & \multicolumn{1}{c}{CMC} & \multicolumn{1}{c}{CAcc} & \multicolumn{1}{c}{CMC} & \multicolumn{1}{c}{CAcc} \\ \hline
\multicolumn{1}{l}{{[}BMP02{]}}  & 42.51\%                          & 20.24\%                           & 44.79\%                          & 19.56\%                            & 38.15\%                         & 20.64\%                           & 49.62\%                          & 22.58\%                           & 31.11\%                          & 10.98\%                           & 62.74\%                          & 21.35\%                           \\ 
\multicolumn{1}{l}{{[}ERB*12{]}} & 24.55\%                         & 11.99\%                           & 30.87\%                          & 9.97\%                            & 35.84\%                          & 17.36\%                           & 32.85\%                          & 13.30\%                           & 27.68\%                          & 10.47\%                           & 55.94\%                         & 13.27\%                           \\ 
\multicolumn{1}{l}{{[}HLJ*15{]}} & 41.78\%                         & 18.51\%                            & 48.41\%                          & 19.90\%                            & 36.89\%                          & 18.29\%                           & 45.33\%                          & 19.66\%                           & 34.85\%                          & 12.78\%                           & 60.72\%                          & 19.48\%                           \\ 
\multicolumn{1}{l}{{[}HKC*18{]}} & 40.20\%                          & 16.72\%                            & 49.39\%                          & 16.29\%                            & 33.63\%                          & 16.32\%                           & 42.87\%                          & 18.30\%                           & 35.95\%                          & 11.85\%                           & 60.55\%                          & 17.87\%                           \\ 
\multicolumn{1}{l}{PCA}          & 20.46\%                          & 9.73\%                            & 30.63\%                          & 9.52\%                            & 30.95\%                          & 16.92\%                           & 26.84\%                          & 10.46\%                           & 26.43\%                          & 9.88\%                           & 49.65\%                          & 11.13\%                           \\ 
\multicolumn{1}{l}{SketchZooms}  & \textbf{62.67\%}                          & \textbf{37.28\%}                            &\textbf{66.44\%}                          & \textbf{31.90\%}                             & \textbf{50.70\%}                          & \textbf{31.49\%}                           & \textbf{57.71\%}                           & \textbf{31.65\%}                           &  \textbf{49.07\%}                          & \textbf{24.39\%}                              & \textbf{71.36\%}                         &  \textbf{27.86\%}                           \\ \hline
                                   & \multicolumn{2}{c}{bumps}                           & \multicolumn{2}{c}{potato chip}                     & \multicolumn{2}{c}{shampoo botle}                   & \multicolumn{2}{c}{waffle iron}                     & \multicolumn{2}{c}{flange}                          & \multicolumn{2}{c}{house}                           \\ \hline
                                   & \multicolumn{1}{c}{CMC} & \multicolumn{1}{c}{CAcc} & \multicolumn{1}{c}{CMC} & \multicolumn{1}{c}{CAcc} & \multicolumn{1}{c}{CMC} & \multicolumn{1}{c}{CAcc} & \multicolumn{1}{c}{CMC} & \multicolumn{1}{c}{CAcc} & \multicolumn{1}{c}{CMC} & \multicolumn{1}{c}{CAcc} & \multicolumn{1}{c}{CMC} & \multicolumn{1}{c}{CAcc} \\ \hline
\multicolumn{1}{l}{{[}BMP02{]}}  & 40.16\%                            &  21.02\%                          & 58.75\%                          & 26.49\%                          & 65.26\%                         & 28.90\%                          & 19.16\%                        & 9.47\%                           & 46.87\%                          & 18.12\%                          &  48.04\%                         & 17.55\%                          \\ 
\multicolumn{1}{l}{{[}ERB*12{]}} & 32.07\%                            &  12.28\%                          & 45.32\%                          & 14.16\%                          & 44.43\%                         & 16.31\%                          & 15.68\%                          & 7.28\%                          & 29.55\%                          & 10.79\%                          &  39.30\%                         & 11.84\%                          \\ 
\multicolumn{1}{l}{{[}HLJ*15{]}} & 39.65\%                            & 18.39\%                           & 60.72\%                          & 24.20\%                          & 63.25\%                         & 23.80\%                          & 19.06\%                       & 10.99\%                          & 40.21\%                          & 16.54\%                          &  44.33\%                         & 14.79\%                           \\ 
\multicolumn{1}{l}{{[}HKC*18{]}} & 36.63\%                            &  15.90\%                           & 55.16\%                          & 19.27\%                          & 62.03\%                          & 23.73\%                          & 18.62\%                         & 10.34\%                          & 38.81\%                         & 13.29\%                          & 40.88\%                         & 11.55\%                          \\ 
\multicolumn{1}{l}{PCA}          & 30.94\%                           &  14.35\%                          & 43.98\%                          & 12.58\%                          & 38.78\%                         & 15.76\%                       & 14.14\%                          & 7.67\%                          & 23.23\%                         & 7.73\%                          &  26.55\%                         & 6.57\%                       \\ 
\multicolumn{1}{l}{SketchZooms}  & \textbf{45.98\%}                           & \textbf{25.40\%}                          &  \textbf{66.63\%}                          & \textbf{30.27\%}                          & \textbf{70.08\%}                         & \textbf{34.73\%}                           & \textbf{23.66\%}                          & \textbf{13.59\%}                         &  \textbf{52.15\%}                         & \textbf{24.93\%}                          & \textbf{51.80\%}                           & \textbf{21.93\%}                          \\ \hline
\end{tabular}
}
\end{table*}

\emph{Competing descriptors}. We compare our method against state-of-the-art descriptors commonly used for local sketch matching tasks, including the radial histograms from ShapeContext \cite{belongie2002shape} and the GALIF descriptor, based on Gabor filters by Eitz~et~al.~\cite{eitz2012sketch}. 
We additionally consider a hand-crafted descriptor consisting on principal component analysis (PCA) over a small neighbourhood of pixels surrounding the target point.  
We also compared SketchZooms against deep learned features. In particular, we considered MatchNet \cite{han2015matchnet}, a patch-based descriptor targeting natural images, and the multiview architecture from Huang. et al. \cite{huang2018learning} to compute local 3D shape descriptors. The latter is closely related to our work, although it relies on a contrastive instead of a triplet loss, and does not apply our hard samples mining strategy during training.
To the best of our knowledge, no deep learning based approaches have been introduced specifically for local sketch matching tasks.
For all the aforementioned methods, we used the official and freely available implementations when possible, or re-implemented them otherwise.
Importantly, for a fair comparison, all deep learning-based methods backbones were adapted to work with AlexNet and retrained with our synthetic sketch dataset.

\emph{Data augmentation and training details.} We computed local descriptors from a set of zoomed sketch views $Z$ (three in our setup) centred on the point of interest $p$.
The network learns rotational invariant descriptors by randomly rotating input images between 0 and 360 degrees with equal probability. 
To keep the descriptor robust to different resolutions, we downsampled the input image size by 30\% and 60\% with a probability of 0.2. 
To diminish sensitivity to the camera-target point distance, we added noise during training to the zoom parameter by sampling camera displacements from a normal distribution (with $\mu=0$ and $\sigma^{2}=0.3$, where 0.3 means 30\% size increment w.r.t.~the original image size). 
Since some views are more densely populated than others, we restrict our training minibatches to have the same number of samples from each view in order to avoid bias. 
Also, since each object class has a different total number of images, we restricted each batch to equally balance the amount of images from each category.
Our data augmentation choices were iterative, and empirically guided by results obtained during the experimentation stage using a validation set.

The network architecture was implemented with PyTorch and trained on NVIDIA Titan Xp GPUs. 
We first initialise the convolutional layers using AlexNet weights trained on the ImageNet data set, as provided in Pytorch. 
The learning rate was set to $l=10^{-5}$ and the network was trained for 185 epochs.
We optimise the objective in Equation~\ref{eq:loss} using Adam optimisation \cite{kingma2014adam} ($\beta_{1}= 0.9, \beta_{2}= 0.999)$ and a batch size of 64 triplets.
We did not use batch normalisation layers or dropout in addition to those already into AlexNet (dropout $p=0.5$ on layer fc6).

\section{Results}\label{sec:results}

\subsection{OpenSketch Benchmark}\label{sec:opensketch}
In order to evaluate our learned features, we conducted a series of comprehensive comparisons against other methods when applied to hand-drawn design sketches. 
We relied on OpenSketch \cite{gryaditskaya2019opensketch}, a dataset of product design sketches containing more than 400 images representing 12 man-made objects drawn by 7 to 15 product designers of varying expertise. 
These design sketches are drawn in a variety of styles and from very different viewports. 
In addition, all images are augmented with a series of corresponding points derived from registered 3D models and manually annotated by designers. 
Importantly, none of the 12 OpenSketch object categories match those in our training dataset: bumps, hairdryer, mixer, potato chip, tubes, waffle iron, flange, house, mouse, shampoo bottle, vacuum cleaner, and  wobble surface. 
This is a key factor to assess the descriptors' generalisation power, particularly those which are learned from our synthetic training data.
Since each line drawing in the dataset has several layers at different progress stages, we filtered them and kept the latest version of the sketch (called \emph{presentation} sketch).
Additionally, since sketches drawn from observation tend to be aligned with the horizontal axis, we altered them by applying a random rotation of $\pm$90$^{\circ}$ to each image before computing the descriptors.
In this way, we effectively evaluate each method's ability to build rotation invariant descriptors.
The correspondences between all image pairs were computed using the Euclidean distance in descriptor space, and choosing the closest target point on each case.

\begin{figure}[!t]
    \centering
    \includegraphics[width=1.0\linewidth]{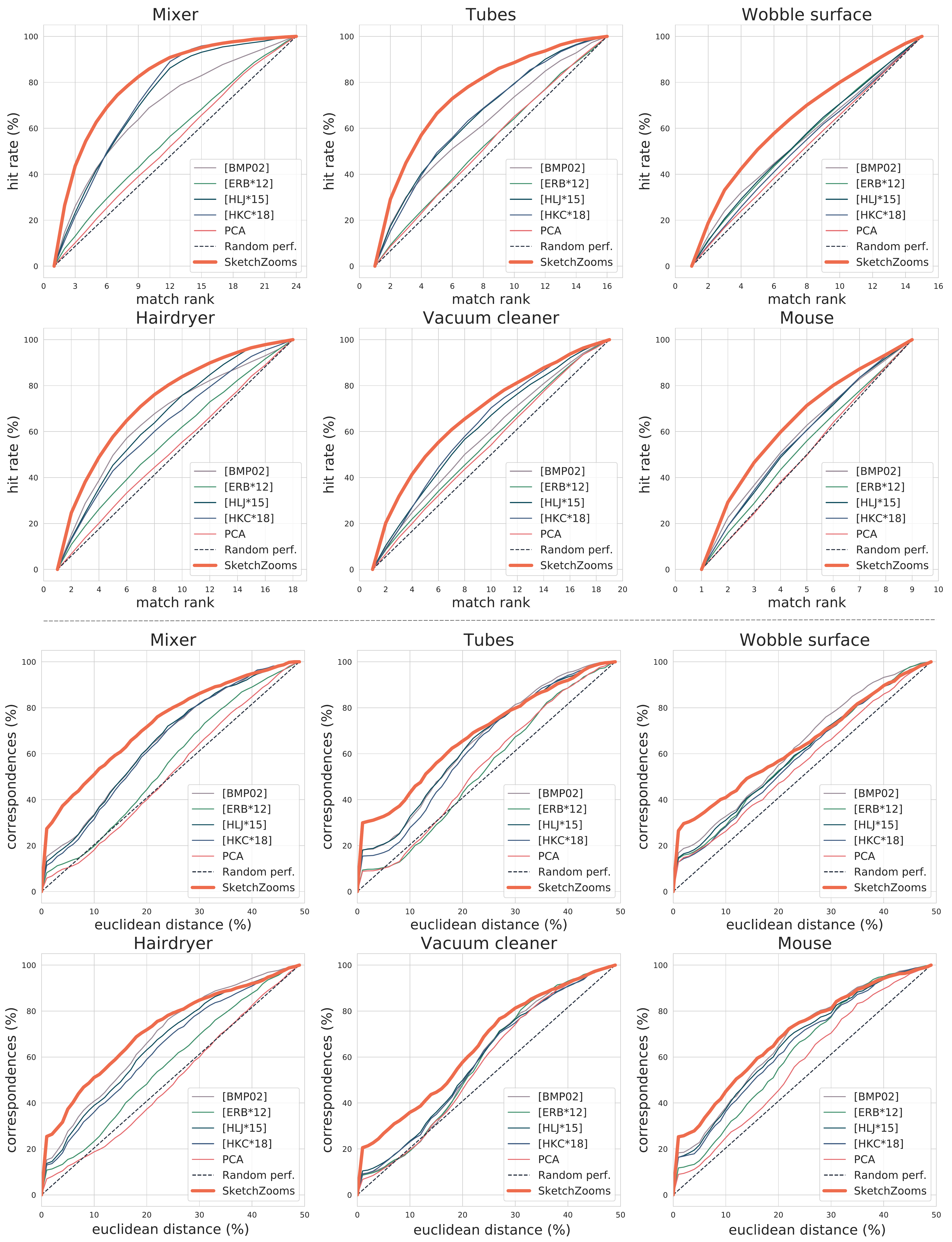}
    \parbox[t]{\columnwidth}{\relax }
    \caption{Top: Cumulative match characteristic plots for the evaluated descriptors on the test data set for mixer, tubes, wobble surface, hairdryer, vacuum cleaner and mouse categories. y-axis accounts for the percentage of matchings retrieved below the raking position indicated on x. Bottom: Correspondence accuracy curves, where x-axis shows normalised Euclidean distance error, and y-axis accounts for the matching percentage of retrieval below the error margin indicated on x.  
    \label{fig:fig_5_eval_metrics}}
\end{figure}

Figure~\ref{fig:fig_4_sparse} shows the retrieved matchings on image pairs from different artists in the dataset for all evaluated methods. 
Overall, our approach was able to successfully exploit the features learned from the synthetic training set when working with hand-drawn images. 
We quantitatively evaluated descriptors on this benchmark by computing the correspondences among all possible pairs of images within each category, a total of 66,320 corresponding points.
Table \ref{tab:cmc-score} reports the performance of the evaluated descriptors over 5 retrieved matches for the CMC and below 5\% normalised Euclidean distance for the CAcc.
We further illustrate these metrics in Figure \ref{fig:fig_5_eval_metrics}.
Our descriptors outperformed the competing methods in both evaluated metrics and across all object categories.
According to the reported metrics, we observed that our learned descriptors outperform the rest, including the patch-based learned descriptors of MatchNet\cite{han2015matchnet} and the multi-view architecture of Huang et al. \cite{huang2018learning}. 
Also, SketchZooms performed better than the hand-engineered local descriptors traditionally used for corresponding line drawings, namely ShapeContext \cite{belongie2002shape} and GALIF \cite{eitz2012humans}.
Following these results, we believe that our method can successfully embed semantically similar feature points in descriptor space closer than other methods, while being stable to changes in view, decorations, and style.  
Moreover, despite the fact that testing categories differ from those used for training, our method can still exploit 3D shape cues to produce fairly general local descriptors that perform favourably compared to general hand-crafted alternatives.

\begin{figure}[!t] 
    \centering
    \includegraphics[width=1.0\linewidth]{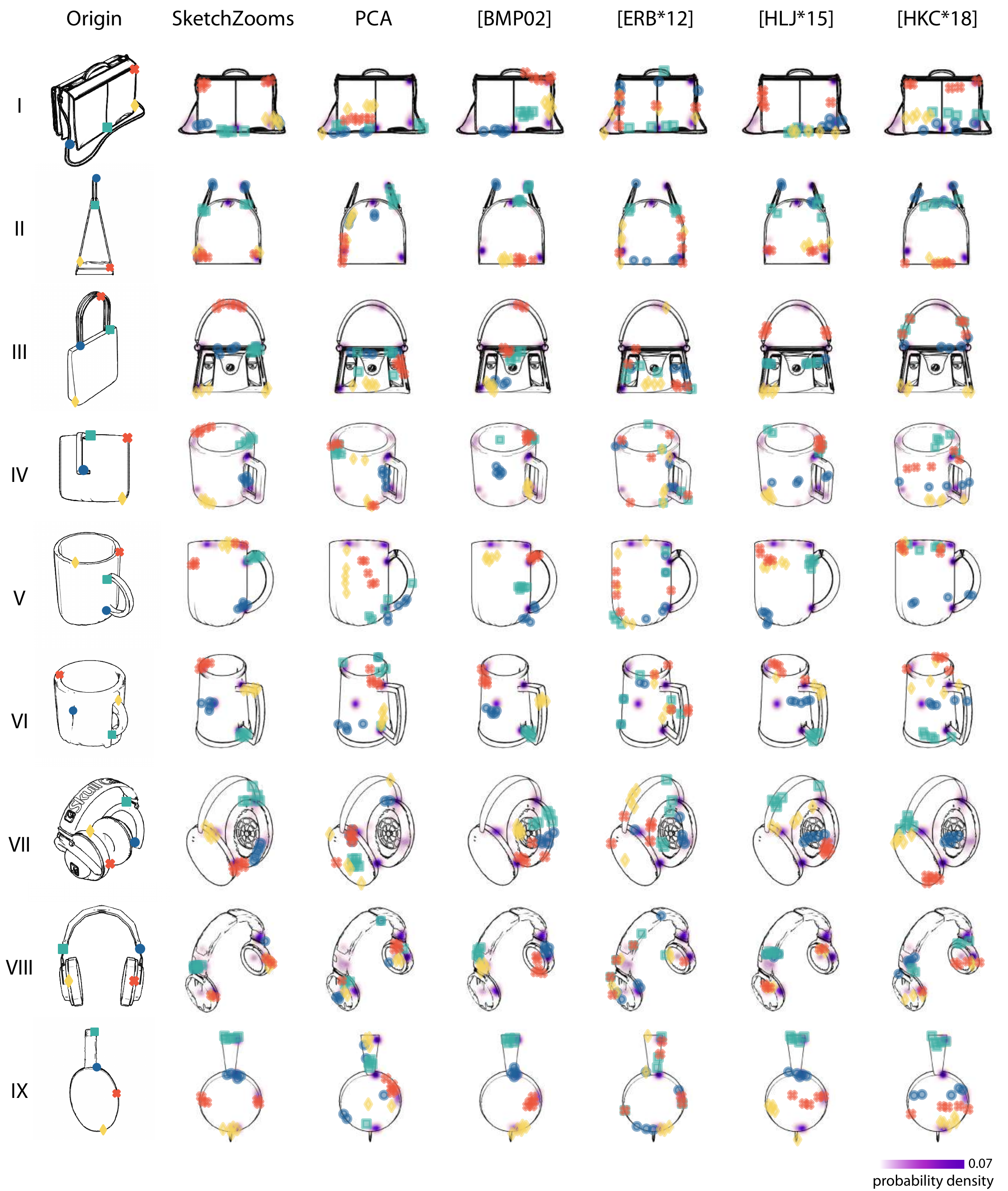}
    \parbox[t]{\columnwidth}{ }
    \caption{Perceptual study. On each row, the first column shows the origin image together with the 4 sampled points shown to participants of the study. All other columns show the top 5 retrieved correspondences computed with each local descriptor among a total of 200 target points. The heatmaps underneath show the probability density distribution of the subjects clicks. The remaining image data from the study is available as supplementary material.
    \label{fig:fig_6_eval_heatmaps}}
\end{figure}

\subsection{Perceptual Study}\label{sec:userstudy}
Humans possess an extraordinary ability to resolve semantic correspondences in multi-view scenarios thanks to their previously-acquired 3D knowledge about the world. 
We conducted a perceptual study to comprehensively assess the relationship between the semantics captured by our descriptors from synthetic data and the decisions made by humans when performing the same matching task on artificial sketches.
Each of the 10 study volunteers was presented with $m=4$ points on a synthetic sketch image (origin), and was instructed to find $m$ corresponding points on a target image. 
We used a total of 40 random image pairs from our synthetic dataset distributed in four categories: bag, chair, earphone, and mug.
Points on the origin images were randomly selected from a larger list of feature points computed over all the study images using the corner detector Good Features to Track~\cite{shi1993good}.
For the target points, we used blue noise sampling to distribute 200 candidate points across the image.
We did not show any of these candidates to the study participants.
Instead, and similar to the approach adopted by BestBuddies~\cite{aberman2018neural}, we registered mouse clicks over targets, and fitted 2D Gaussian distributions over the coordinates annotated by the users. 
Overall, we observe all participants consistently corresponded points on target images within specific regions.
After the subjects sessions, we retrieved matchings among origin and target images by selecting the closest points in the Euclidean space for all compared methods.
We then defined a similarity measure by evaluating the average fitted probability density function on the top 5 retrieved matches for each query point. 
Higher similarity scores are then assigned to regions where the consensus among users and the automatically retrieved points is strong, and vice-versa. 
We averaged the scores for all the points within each object category and summarised the results in Table~\ref{tab:user-score}.

Figure~\ref{fig:fig_6_eval_heatmaps} illustrates matchings computed with each local descriptor and the areas where the subjects consensus was stronger.
When the participants had to disambiguate between points with identical semantics in symmetric views of an object (Figure~\ref{fig:fig_6_eval_heatmaps} II, III and IX), most of them decided to choose those on the same relative position with respect to its counterpart in the origin image. SketchZooms descriptors often find multiple semantically similar candidates on both sides of the vertical symmetry plane (such in rows II, III and IV from Figure~\ref{fig:fig_6_eval_heatmaps}).
This aspect of our descriptors make them more robust to arbitrary rotations and reflections, as shown in Section~\ref{sec:opensketch}. 
An extended discussion about symmetry is presented in Section~\ref{sec:limit}.

Correspondences obtained with our descriptors are closer to hot areas than those produced with other methods.
ShapeContext also produced accurate descriptors for these images. However, this is contradictory with the performance previously observed on the OpenSketch benchmark. We believe this difference is likely due to ShapeContext lacking a learning strategy, which renders a method unable to generalize to more complex, realistic sketch data such as OpenSketch.
ShapeContext is a hand-crafted descriptor designed to correspond shape outlines that look similar and clean, like the synthetic sketches used in the study. 
When corresponding hand-drawn images with severe projection distortions, multiple rough strokes and shading, ShapeContext fails to recognise the underlying similar local shapes (Figure~\ref{fig:fig_4_sparse}). 
The full set of images from subjects data is available as supplementary material. 

\begin{table}[t]
\caption{Perceptual study metrics.}
\resizebox{\columnwidth}{!}{%
\begin{tabular}{l c c c c}
\hline
                                   & bag                                       & chair                                     & earphone                                  & mug                                       \\ \hline
\multicolumn{1}{l}{{[}BMP02{]}}  & 0.023$\pm$0.013                           & 0.024$\pm$0.008                           & 0.021$\pm$0.011                           & 0.022$\pm$0.010                           \\ 
\multicolumn{1}{l}{{[}ERB*12{]}} & 0.005$\pm$0.003                           & 0.006$\pm$0.004                           & 0.009$\pm$0.006                           & 0.008$\pm$0.004                           \\ 
\multicolumn{1}{l}{{[}HLJ*15{]}} & 0.014$\pm$0.010                           & 0.019$\pm$0.005                           & 0.016$\pm$0.006                           & 0.018$\pm$0.010                           \\ 
\multicolumn{1}{l}{{[}HKC*18{]}} & 0.011$\pm$0.005                           & 0.015$\pm$0.008                           & 0.014$\pm$0.003                           & 0.013$\pm$0.007                           \\ 
\multicolumn{1}{l}{PCA}          & 0.003$\pm$0.002                           & 0.006$\pm$0.004                           & 0.006$\pm$0.003                           & 0.004$\pm$0.004                           \\ 
\multicolumn{1}{l}{SketchZooms}  & \textbf{0.025$\pm$0.007} & \textbf{0.025$\pm$0.009} & \textbf{0.024$\pm$0.009} & \textbf{0.024$\pm$0.012} \\ \hline
\end{tabular}
}
 \label{tab:user-score}
\end{table}

\begin{table*}[t]
\centering
\caption{Ablation study metrics for the OpenSketch benchmark. Cumulative match characteristic (top 5 rank) and correspondence accuracy for normalised Euclidean distance at 5\% on all evaluated backbone architectures.}
\label{tab:ablation}
\resizebox{\textwidth}{!}{%
\begin{tabular}{lcccccccccccc}
\hline
          & \multicolumn{2}{c}{mixer}           & \multicolumn{2}{c}{tubes}           & \multicolumn{2}{c}{wobble surface}  & \multicolumn{2}{c}{hairdryer}       & \multicolumn{2}{c}{vacuum cleaner}  & \multicolumn{2}{c}{mouse}           \\ \hline
          & CMC              & CAcc             & CMC              & CAcc             & CMC              & CAcc             & CMC              & CAcc             & CMC              & CAcc             & CMC              & CAcc             \\ \hline
ResNet-18 & 60.98\%          & 33.83\%          & 63.99\%          & 29.89\%          & 47.89\%          & 29.97\%          & 51.43\%          & 27.27\%          & 45.25\%          & 20.85\%          & 68.82\%          & 23.82\%          \\
VGG-19    & \textbf{63.95\%} & 37.15\%          & \textbf{69.66\%} & \textbf{36.16\%} & 49.60\%          & 30.92\%          & \textbf{57.91\%} & 31.59\%          & 48.62\%          & 23.11\%          & \textbf{72.68\%} & 27.58\%          \\
AlexNet   & 62.67\%          & 37.28\%          & 66.44\%          & 31.90\%          & \textbf{50.70\%} & \textbf{31.49\%} & 57.71\%          & \textbf{31.65\%} & \textbf{49.07\%} & \textbf{24.39\%} & 71.36\%          & \textbf{27.86\%} \\ \hline
          & \multicolumn{2}{c}{bumps}           & \multicolumn{2}{c}{potato chip}     & \multicolumn{2}{c}{shampoo botle}   & \multicolumn{2}{c}{waffle iron}     & \multicolumn{2}{c}{flange}          & \multicolumn{2}{c}{house}           \\ \hline
          & CMC              & CAcc             & CMC              & CAcc             & CMC              & CAcc             & CMC              & CAcc             & CMC              & CAcc             & CMC              & CAcc             \\ \hline
ResNet-18 & 43.16\%          & 21.54\%          & 65.64\%          & 28.20\%          & 71.36\%          & 32.95\%          & 23.40\%          & 13.52\%          & 51.44\%          & 21.54\%          & 51.32\%          & 19.32\%          \\
VGG-19    & 44.97\%          & \textbf{25.43\%} & \textbf{71.08\%} & \textbf{32.92\%} & \textbf{72.10\%} & 34.17\%          & \textbf{24.49\%} & \textbf{14.65\%} & \textbf{52.73\%} & 23.57\%          & \textbf{55.76\%} & \textbf{24.26\%} \\
AlexNet   & \textbf{45.98\%} & 25.40\%          & 66.63\%          & 30.27\%          & 70.08\%          & \textbf{34.73\%} & 23.66\%          & 13.59\%          & 52.15\%          & \textbf{24.93\%} & 51.80\%          & 21.93\%         \\ \hline
\end{tabular}
}
\end{table*}

\subsection{Triplet vs. Contrastive Loss}
Similar to our approach, the work by Huang~et~al.~\cite{huang2018learning} relies on a multi-view architecture to learn descriptors for 3D models, trained using a contrastive loss, and random minibatches built during the learning phase without using any sampling heuristic.
On the other hand, our work relies on a triplet loss function and uses a custom training schedule that can potentially benefit other application dealing with unbalanced sets of views.
In order to empirically compare these two approaches, we trained an adaptation of the approach of Huang~et~al.~to this specific problem, keeping all training hyperparameters and the aforementioned data augmentation strategies to avoid mixing effects in the evaluations.
SketchZooms performed better on all testing categories (Table \ref{tab:cmc-score}), with average CMC = 55.52\% (SketchZooms) over CMC = 42.89\% (\cite{huang2018learning}), and CAcc = 27.95\% (SketchZooms) over CAcc = 17.68\% (\cite{huang2018learning}). 
These experiments support our hypothesis that a combined training strategy based on a triplet loss and a smart data sampling procedure is of paramount importance in order to improve results with respect to basic contrastive losses and random samplings.
Additionally, results indicate that multi-view convolutional neural network architectures can learn meaningful semantic descriptors in contexts where the texture image information is very scarce and ambiguous, like line drawings.

\subsection{Architecture Alternatives}\label{sec:abblation}
We further investigated the effect of adopting other network architecture as backbones in our pipeline. 
Therefore, we trained our framework using two alternative architectures, namely VGG19 \cite{simonyan2014very} (133,387,752 total parameters) and ResNet-18 \cite{he2016deep} (11,242,176 total parameters). 
Both models were fine-tuned from ImageNet weights using the same hyperparameter settings as AlexNet (40,796,610 total parameters), with the only exception of the batch size for VGG19, which had to be reduced by half due to the large memory requirements of the network.
Table~\ref{tab:ablation} summarises the performance over the OpenSketch benchmark data.
Overall, we found a slight improvement on the evaluated metrics across most object categories when using VGG19 architecture. 
We believe this is likely due to the well-known properties of VGG19 as a feature extractor, observed in multiple different applications~\cite{sharif2014cnn}.
It must be noticed, however, that these advantages come at the cost of a much slower training due to the significant amount of parameters on this network, most of them originated in the last series of fully connected layers.
ResNet-18, on the other hand, performed much more poorly in our experiments, probably due to the lack of a stack of fully connected layers and the usage of global average pooling.

\section{Robustness and Limitations}\label{sec:limit}
We now discuss the overall behavior of our method under challenging scenarios and its main limitations. 

\begin{figure}[!t] 
    \centering
    \includegraphics[width=1.0\linewidth]{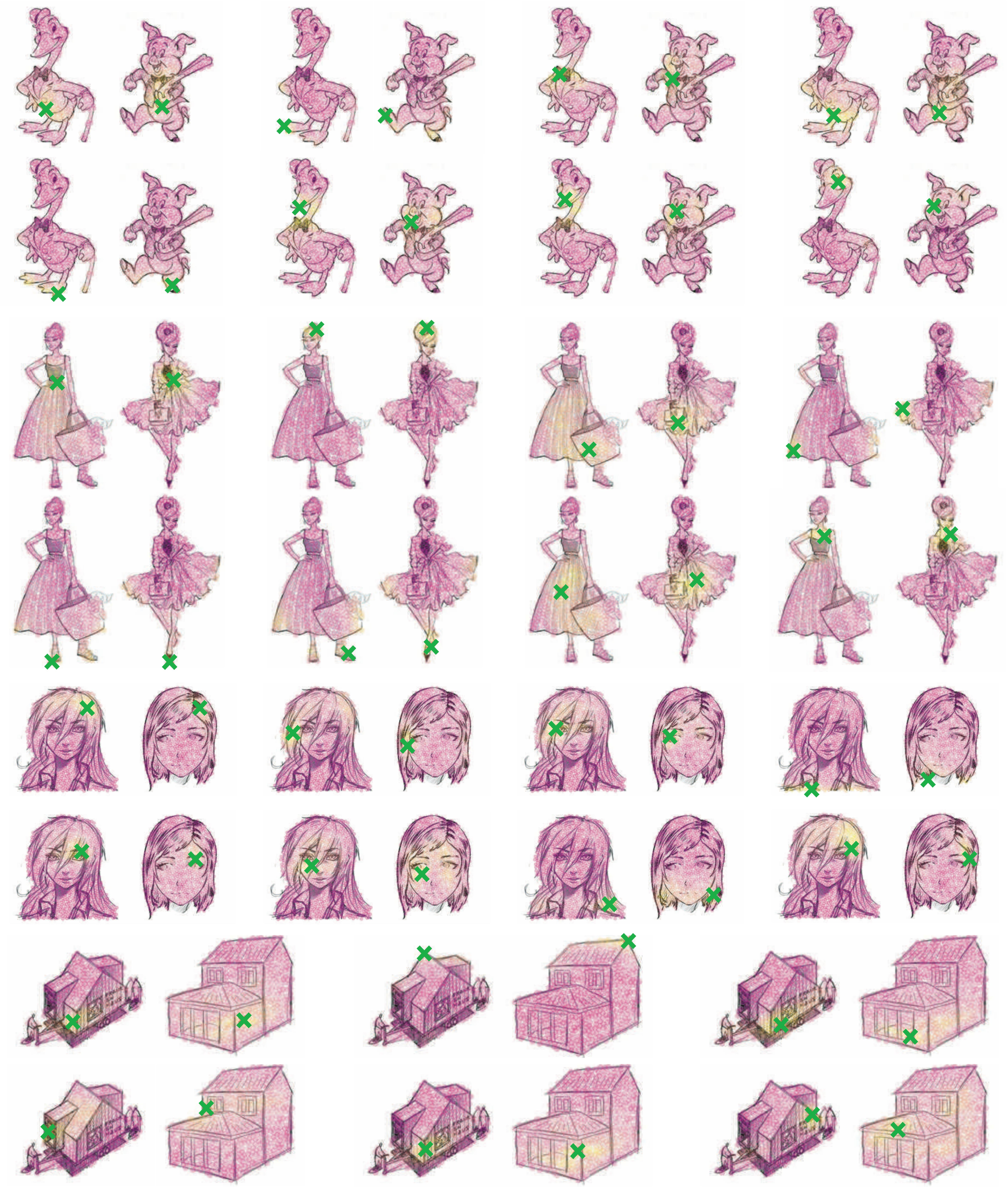}
    \parbox[t]{\columnwidth}{}
    \caption{SketchZooms dense correspondences on line drawings from different styles (from top to bottom: cartoons, fashion, manga, and architectural sketches). Colors indicate distance to target point (green cross) in feature space. For each line drawing, between 400 and 1000 points were randomly sampled and corresponded using Euclidean distance.}
    \label{fig:fig_7_eval_dense}
\end{figure}

\emph{Robustness to sketchiness}. Adopting \emph{Apparent Ridges} as our dataset rendering engine allowed our method to be robust to typical drawings' sketchiness.
Synthetic images rendered with this method often contain wiggly lines and several other imperfections. 
Our experimental setup allowed us to evaluate SketchZooms ability to deal with very different drawing styles (Figure~\ref{fig:fig_4_sparse}, V and VII), even with overlayed construction lines and shadows (Figure~\ref{fig:fig_4_sparse}, IX and XI).
However, extremely rough drawings, with an excessive amount of construction lines or extreme lighting can harm the performance of our descriptors (i.e. cross-hatched areas in houses from Figure~\ref{fig:fig_7_eval_dense}).

\emph{Symmetry}. Most man-made objects are symmetric with respect to at least one plane in 3D space. 
Our features are strongly biased by the image semantic information, and sometimes can mismatch on symmetric points on the target sketch (see Figure~\ref{fig:fig_6_eval_heatmaps}).
Symmetry mismatches happen more often when trying to correspond extreme viewports, like the side and front of two target objects. 
For all results reported in this paper, we used the Euclidean distance in feature space to retrieve correspondences and compute metrics.
However, simultaneously matching several points can help disambiguate these symmetries --i.e.~combinatorial optimisation methods like the Hungarian algorithm~\cite{kuhn1955hungarian} can help refine more coherent matchings than using a simple strategy of matching closer points in descriptor space. 
An interesting future research direction is to explore ways to incorporate orientation tags in the training phase or to involve users actively in refining correspondences on the fly. 

\begin{table}[t]
\caption{Descriptors performance under random zoom and rotations up to the values indicated in top rows for each table. Cumulative match characteristic is reported for the top 5 rank and correspondence accuracy for normalised Euclidean distance at 5\% of the image width.}
\label{tab:zooms}
\resizebox{\columnwidth}{!}{%
\begin{tabular}{lllllll}
\hline
max. zoom         & \multicolumn{2}{c}{$\pm$10\%}                        & \multicolumn{2}{c}{$\pm$20\%}                        & \multicolumn{2}{c}{$\pm$40\%}                        \\ \hline
             & \multicolumn{1}{c}{CMC} & \multicolumn{1}{c}{CAcc} & \multicolumn{1}{c}{CMC} & \multicolumn{1}{c}{CAcc} & \multicolumn{1}{c}{CMC} & \multicolumn{1}{c}{CAcc} \\ \hline
{[}BMP02{]}  & 46.69\%                 & 20.44\%                  & 44.85\%                 & 19.38\%                  & 44.73\%                 & 20.38\%                  \\
{[}ERB*12{]} & 34.31\%                 & 12.61\%                  & 33.75\%                 & 12.31\%                  & 34.02\%                 & 13.17\%                  \\
{[}HLJ*15{]} & 45.42\%                 & 18.67\%                  & 44.16\%                 & 17.56\%                  & 43.18\%                 & 18.07\%                  \\
{[}HKC*18{]} & 43.04\%                 & 16.82\%                  & 41.68\%                 & 15.94\%                  & 42.43\%                 & 16.95\%                  \\
PCA          & 30.09\%                 & 11.07\%                  & 30.43\%                 & 10.91\%                  & 30.21\%                 & 11.93\%                  \\
SketchZooms  & \textbf{55.39\%}        & \textbf{27.52}           & \textbf{53.97\%}        & \textbf{26.69\%}         & \textbf{52.93\%}        & \textbf{26.78\%}         \\ \hline
          
max. rotation     & \multicolumn{2}{c}{$\pm$45$^{\circ}$}                         & \multicolumn{2}{c}{$\pm$90$^{\circ}$}                         & \multicolumn{2}{c}{$\pm$180$^{\circ}$}                        \\ \hline
             & \multicolumn{1}{c}{CMC} & \multicolumn{1}{c}{CAcc} & \multicolumn{1}{c}{CMC} & \multicolumn{1}{c}{CAcc} & \multicolumn{1}{c}{CMC} & \multicolumn{1}{c}{CAcc} \\ \hline
{[}BMP02{]}  & 51.53\%                 & 24.79\%                  & 45.60\%                 & 19.74\%                  & 38.06\%                        & 15.10\%                          \\
{[}ERB*12{]} & 36.79\%                 & 14.27\%                  & 34.51\%                 & 12.42\%                  & 31.58\%                        & 10.97\%                          \\
{[}HLJ*15{]} & 49.37\%                 & 21.22\%                  & 44.51\%                 & 18.11\%                  & 35.83\%                        & 16.66\%                          \\
{[}HKC*18{]} & 54.78\%                 & 17.39\%                  & 42.89\%                 & 15.95\%                  & 34.91\%                        & 12.52\%                          \\
PCA          & 30.23\%                 & 11.32\%                  & 30.21\%                 & 11.03\%                  & 30.19\%                     & 10.90\%                          \\
SketchZooms  & \textbf{59.38\%}        & \textbf{31.14\%}         & \textbf{55.69\%}        & \textbf{27.95\%}         & \textbf{44.58\%}                       & \textbf{20.72\%}                          \\ \hline
\end{tabular}
}
\end{table}

\emph{Zoom and rotation sensitivity}.
As mentioned in Section~\ref{sec:approach}, in order to compute a descriptor for a given image point, we need to successively crop three zoomed images surrounding it.
These images are aggregated and transformed by the SketchZooms network to produce a descriptor of the point. 
We pick the zoom parameter value in order to include some information of the strokes composing the target image, since providing three empty images to the network would produce undesired outputs. 
In particular, for all results presented in this paper we fixed zoomed images sides to be 10\%, 20\%, and 40\% of the total image length (512 pixels in our experiments).
In general, OpenSketch images are relatively on the same scale, occupying at least two thirds of the total width and aligned with the horizontal image plane. 
To assess the effect of different zooms and rotations, we performed a controlled study where the testing images were zoomed in or out at different scales before computing the descriptors.
In particular, we segmented the objects from OpenSketch images and re-scaled them randomly at different maximum sizes $\pm$10\%, $\pm$20\%, and $\pm$40\%.
We measured size as the maximum distance among all pairs of stroke pixels for each image.  
We also generated versions of the dataset where images were randomly rotated up to $\pm$45$^{\circ}$, $\pm$90$^{\circ}$, and $\pm$180$^{\circ}$.
Then, we computed the evaluated metrics on all possible corresponding pairs within each category. 
Table~\ref{tab:zooms} summarises the results.
While SketchZooms performance is not greatly affected by these parameters, zooming too much can lead to cases in which the three cropped images have any stroke information, while zooming too little could miss details, degrading the output descriptor quality. 

\begin{figure}[t] 
    \centering
    \includegraphics[width=1.0\linewidth]{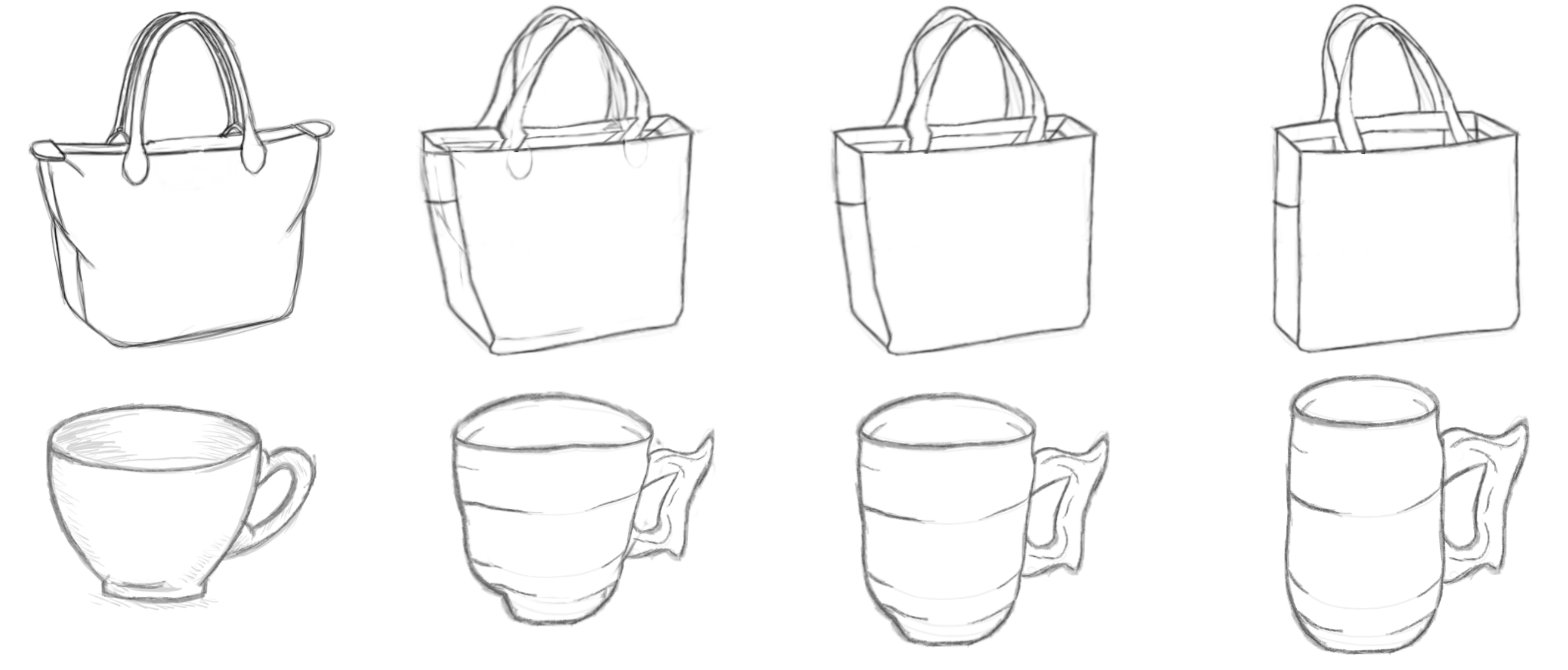}
    \parbox[t]{\columnwidth}{ }
    \caption{Image morphing sequences using SketchZooms descriptors for corresponding two target sketches. A non-linear alpha blending map was computed from point distances in the SketchZooms feature space. 
    \label{fig:fig_8_apps_morphing}}
\end{figure}

\emph{Generalisation to unseen line drawing styles}. Finally, we show the capability of SketchZooms to perform on images significantly different than the ones used for training. 
We selected pairs of sketches from the Yan et al. \cite{yan2020benchmark} public dataset and computed dense correspondences.
Figure~\ref{fig:fig_7_eval_dense} shows exemplary outputs of corresponding points in cartoons, manga, fashion, and architectural sketches. 
Overall, our learned features produced plausible matchings. 
Importantly, the distance field in feature space reveals a smooth embedding, where semantically and geometrically similar points are close to each other. 
This smoothness does not appear to be significantly altered by the rough shading variation and other discontinuities in the images.
Even if none of these sketch categories were used to train our model, our highly diverse synthetic dataset used for training ensured a regularisation effect, allowing generalisation to unseen styles.

\section{Applications}\label{sec:applications}

\emph{Image morphing for shape exploration.} 
Inspired by the recent work of Arora~et~al.~\cite{arora2017sketchsoup}, we implemented an image morphing algorithm based on the image mapping obtained from the SketchZooms features. 
The goal is to allow exploration of the continuous design space between two sketches while smoothing views and shape transitions. 
We start by computing motion paths between sparse SketchZooms corresponding points, and then interpolate them into dense smooth trajectories. 
We sample $k=10$ correspondences evenly distributed over the input-target pair. 
Then, we compute a Delaunay triangulation of the image space using the sampled points as input. 
For each triangle, we estimate an affine transformation that maps both triangulations on a number of steps $\mathbf{s}=50$.
We implemented a non-linear alpha blending function to reduce ghosting effects for a pixel $p$ at a step $\mathbf{s}$ defined as:
\begin{equation}
\alpha_{p}\left(\mathbf{s}\right) = \frac{1}{2}+ \frac{1}{2}\tanh \left( \dfrac{\mathbf{s}-\delta(p)}{ \rho(p)} \right),
\end{equation}
where $\delta$ and $ \rho$ are linear functions of the pixel confidence score to keep the sigmoid outputs in the $[0,1]$ interval. 
This blending function ensures that well matched regions smoothly transition into other images, while regions with poor matching disappear quickly from the image (Figure~\ref{fig:fig_8_apps_morphing}).

\begin{figure}[t] 
    \centering
    \includegraphics[width=1.0\linewidth]{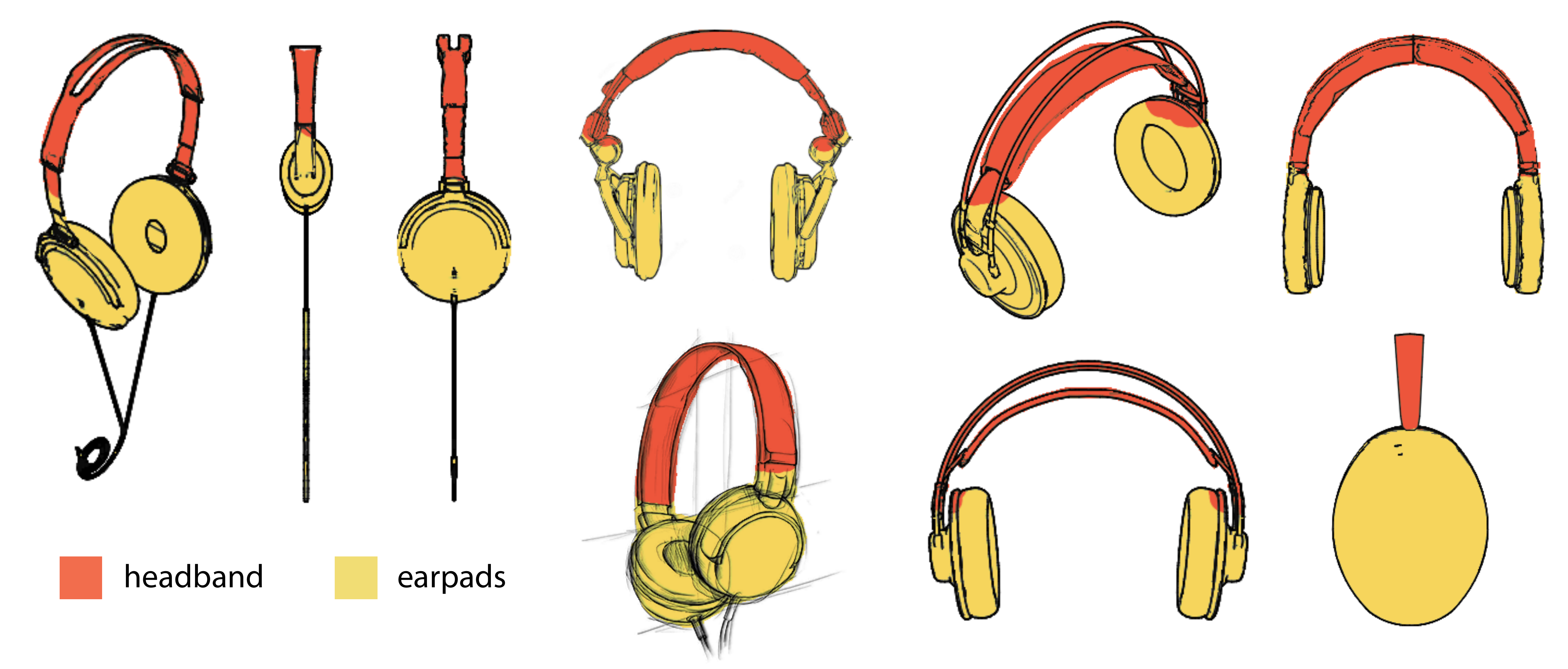}
    \parbox[t]{\columnwidth}{ }
    \caption{SketchZooms features re-purposed for semantic segmentation. Labels can be used to decompose sketches into different layers or as pixel-wise semantic tags for coloring.  
    \label{fig:fig_9_apps_layering}}
\end{figure}

\emph{Part segmentation.}  
Sketch segmentation has been addressed before as an instance of colorization~\cite{sykora2009lazybrush} and simplification~\cite{noris2012smart,liu2018strokeaggregator}. 
Segmentation can be used for different applications, like adding depth information to line drawings or applying global illumination effects~\cite{sykora2010adding, sykora2014ink}. 
Similarly, SketchZooms' features can be used to perform automatic semantic layering and coloring, since painting has much in common with image segmentation.
Specifically, we first manually segmented hand-drawn images from the headphone category (10 in our test application).
Then, we computed SketchZooms' features for a subset of 2D points on every sketch using blue noise sampling, and used them to train a \emph{C-SVM} classifier which learns to predict labels from our descriptors. 
Formally, we solve for: 
\begin{equation}
\begin{array}{c}{\min_{w, b, \zeta} \mathlarger {\frac{1}{2} w^{T} w+C \sum_{i=1}^{n} \zeta_{i}}}
\\ {\text { subject to } y_{i}\left(w^{T} \phi\left(x_{i}\right)+b\right) \geq 1-\zeta_{i}}
\\ {\zeta_{i} \geq 0, i=1, \ldots, n}\end{array},
\end{equation}
where $C$ is the capacity constant (set to $C=1$), $\boldsymbol{w}$ is the vector of coefficients, and $\boldsymbol{\zeta_{i}}$ represents parameters for handling non-separable data.
The index $i$ labels the $n$ training cases ($n=2$ in our setup).
Figure~\ref{fig:fig_9_apps_layering} shows the semantic segmentations obtained with our classifier.

\begin{figure}[t]
    \centering
    \includegraphics[width=1.0\linewidth]{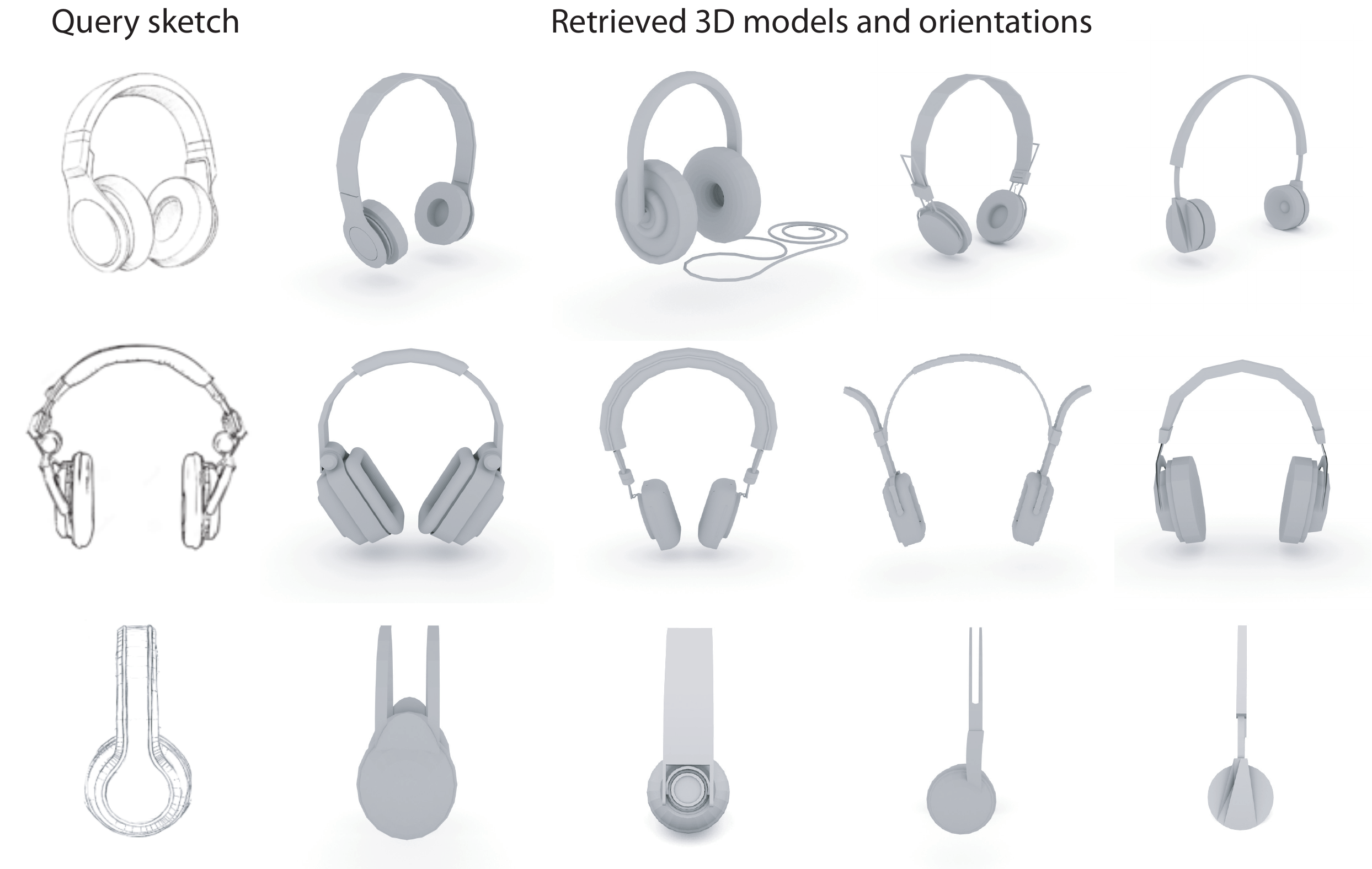}
    \parbox[t]{\columnwidth}{ }
    \caption{Results from our 3D shape search engine. Even though the searched models had no ground truth correspondence on the model database, our algorithm returned plausible shapes. Our features additionally provide information about the sketch view, allowing to automatically orient models to the query sketch.
     \label{fig:fig_10_apps_shape_ret}}
\end{figure}

\emph{Sketch-based 3D shape retrieval.} As shown in Section~\ref{sec:related}, much of the work on image features for sketches was proposed in the context of 3D retrieval applications. 
In order to test the potential of our features in this task, we implemented a 3D model search engine based on our local descriptors.
We computed SketchZooms descriptors for 70 random point samples over the synthetic line drawings of 70 earphone models (4900 points in total distributed among 3 different viewports).
At searching time, we sample 1000 points from query sketches using blue noise sampling, and retrieve the candidate model list using L2 distance w.r.t.~query points. 
This simple strategy retrieves similar models in the database (Figure~\ref{fig:fig_10_apps_shape_ret}).
Additionally, our search engine can accurately determine which camera viewport best matches the query sketch in order to consistently orient 3D models, demonstrating the capability of our feature vectors to encode viewport information.

\section{Conclusions}
We presented SketchZooms, a learnable image descriptor for corresponding sketches.
To the best of our knowledge, SketchZooms is the first data-driven approach that automatically learns semantically coherent descriptors to match sketches in a multi-view context. 
Aiming this with deep neural networks was unfeasible before due to data limitation, as massively collecting sketches from artists and designers is extremely challenging. 
We have put together a vast collection of synthetic line drawings from four human-made objects categories and camera viewports commonly adopted by designers. 
This dataset can be easily extended with our pipeline as more 3D models become available.
More importantly, our learned features were able to generalise to sketches in the wild directly from the synthetic data. 

Our results offer interesting future directions of research. 
Apart from the already mentioned applications, like 3D part segmentation, semantic morphing and sketch-based retrieval, more technical research venues are also raised by this proposal.
It is relevant to investigate whether other viewport configurations are possible without introducing much ambiguity into the descriptor space. 
Also, recent approaches have proposed to use semi-supervised hand-drawn images to improve network performance~\cite{simo2018mastering}.
Investigating whether explicit treatment of domain shifts can boost performance on our hand-drawn data set is an interesting future direction to explore.
Finally, a deep study on how humans perform matching tasks on the sketch image domain would be very beneficial to build more accurate descriptors.

\section*{Acknowledgements}
This study was supported by Agencia Nacional de Promoción Científica y Tecnológica, Argentina, PICT 2018-04517, PID UTN 2018 (SIUTNBA0005139), PID UTN 2019 (SIUTNBA0005534), and NVIDIA GPU hardware grant that provided two Titan Xp graphic cards. We also want to deeply acknowledge the short research internship program from Universidad Nacional de la Patagonia San Juan Bosco, Argentina, and all the perceptual study volunteers.

\bibliographystyle{eg-alpha-doi} 
\bibliography{references}       

\end{document}